\documentclass{llncs}
\usepackage{graphicx}
\usepackage{subfigure}
\usepackage{times}
\usepackage{xcolor}
\usepackage{soul}
\usepackage[utf8]{inputenc}
\usepackage[small]{caption}
\usepackage[hidelinks]{hyperref}

\usepackage{graphicx}
\usepackage{float}

\title{A Virtual Testbed for Critical Incident Investigation with Autonomous Remote Aerial Vehicle Surveying, Artificial Intelligence, and Decision Support}
\titlerunning{A Virtual Testbed for Critical Incident Investigation}

\author{
David L. Smyth,
Sai Abinesh,
Nazli B. Karimi,\\
Brett Drury,
Ihsan Ullah,
Frank G. Glavin,
Michael G. Madden}
\authorrunning{Ihsan Ullah et al.}

\institute{School of Computer Science, National University of Ireland Galway, Ireland
\email{michael.madden@nuigalway.ie}}
\begin{document}

\maketitle

\begin{abstract}
Autonomous robotics and artificial intelligence techniques can be used to support human personnel in the event of critical incidents. These incidents can pose great danger to human life. Some examples of such assistance include: multi-robot surveying of the scene; collection of sensor data and scene imagery, real-time risk assessment and analysis; object identification and anomaly detection; and retrieval of relevant supporting documentation such as standard operating procedures (SOPs). These incidents, although often rare, can involve chemical, biological, radiological/nuclear or explosive (CBRNE) substances and can be of high consequence. Real-world training and deployment of these systems can be costly and sometimes not feasible. For this reason, we have developed a realistic 3D model of a CBRNE scenario to act as a testbed for an initial set of assisting AI tools that we have developed.\footnote{This research has received funding from the European Union's Horizon 2020 Programme under grant agreement No. 700264.}
\end{abstract}

\section{Background and Related Research}
We have developed a bespoke virtual environment (VE)  model of a critical incident using a state-of-the-art games engine. We use this model to test a range of assisting AI technologies related to information gathering, real-time analytics and decision support.\\
\indent We developed the VE with the core purpose of using it as a testbed for the development of a range of investigation assisting AI tools. VEs have also been used to train first responder personnel in near photo-realistic yet safe conditions. Chroust and Aumayr \cite{chroust2017resilience} note that virtual reality can support training by allowing simulations of potential incidents, as well as the consequences of various courses of action, in a realistic way. There are virtual reality training systems which solely focus on CBRN disaster preparedness. Some of these are outlined by Mossel \emph{et al.} \cite{mossel2017requirements}. Other example uses of virtual worlds include \emph{Second Life} and \emph{Open Simulator} \cite{cohen2013emergency,cohen2013tactical}.\\
\indent CBRNE incident assessment is a critical task which poses significant risks and endangers the lives of human investigators. For this reason, many research projects focus on the use of robots such as Micro Unmanned Aerial Vehicles (MUAV) to carry out remote sensing in such hazardous environments \cite{marques2017gammaex,baums2017response}. Others can include CBRNE mapping for first responders \cite{jasiobedzki2009c2sm} and multi-robot reconnaissance  for detection of threats \cite{schneider2012unmanned}.

\section{A Virtual Testbed for Critical Incidents}
We have developed and implemented a baseline set of decision support systems for investigating critical incidents. In order to test these in an efficient and cost effective manner, we have developed 3D world models of typical CBRNE incidents using a physics-based game engine. These models include virtual representations of Robotic Aerial Vehicles (RAVs).\\
\indent After identifying the area of interest, multiple RAVs are deployed to survey the scene. The RAVs, which are fitted with sensors and cameras, operate as a multi-agent robot swarm and divide the work up between them. All information is relayed to a central hub in which our Image Analysis module uses a Deep Neural Network (DNN) to detect and identify relevant objects in images taken by RAV cameras. It also uses a DNN to perform pixel-level semantic annotation of the terrain, to support subsequent route-planning for Robotic Ground-based Vehicles (RGVs). Our Probabilistic Reasoning module assesses the likelihood of different threats, as information arrives from the scene commander, survey images and sensor readings. Our Information Retrieval module ranks documentation, using TF-IDF, by relevance to the incident. All interactions are managed by our purpose-built JSON-based communications protocol, which is also supported by real-world RAVs, cameras and sensor systems. This keeps the system loosely coupled, and will support future testing in real-world environments.\\
\indent This work was undertaken as part of a project called ROCSAFE (Remotely Operated CBRNE Scene Assessment and Forensic Examination) and this demonstration overview is based on Smyth \emph{et al.} \cite{smyth2018virtual}. 

\subsection{Modelling a Critical Incident Scenario}
To facilitate the development and testing of our AI tools, we have designed, developed and publicly released a VE \cite{repo} using the \emph{Unreal Engine} (UE). This is a suite of tools for creating photo-realistic simulations with accurate real-world physics. UE is open source, scalable and supports plugins that allow the integration of RAVs and RGVs into the environment. For this demonstration, we chose an operational scenario to model that consists of a train carrying radioactive material in a rural setting. We used Microsoft's \emph{AirSim} \cite{airsim2017fsr} plugin to model the RAVs. AirSim exposes various APIs to allow fine-grain control of RAVs, RGVs and their associated components. We have replicated a number of APIs from real-world RAV and RGV systems to facilitate the application of our AI tools to real-world critical incident use-cases in the future, after firstly testing them in the VE. 
\subsection{Communications}
A secure purpose-built JSON-format protocol was developed for the communications between subsystems. We used a RESTful API because of the fewer number of messages at pre-defined intervals \cite{Richardson:2013:RWA:2566876}.
The communication protocol not only provides autonomy to several vehicles but it is also flexible enough to integrate with various components using different standards, protocols and data types. In this demonstration, we concentrate on RAVs. Since decision making may happen within each RAV’s single-board computer, we have also facilitated direct communication between the RAVs.
\subsection{Autonomous Surveying and Image Collection}
Our multi-agent system supports the autonomous mapping of the virtual environment. It involves discretizing a rectangular region of interest into a set of grid points. At each point, the RAV records a number of images and metadata. Four bounding GPS coordinates (corner points of a rectangle) can be passed in through a web-based user interface. 
\\ \indent Our planning algorithm develops agent routes at a centralized source and distributes the planned routes to each agent in the multi-agent system \cite{gem}. A greedy algorithm is used in the current implementation to generate subsequent points in each agent’s path by minimizing the distance each agent needs to travel to an unvisited grid point. Current state-of-the-art multi-agent routing algorithms use hyper-heuristics, which out-perform algorithms that use any individual heuristic \cite{hyper}. We intend to integrate this approach with learning algorithms such as Markov Decision Processes \cite{ulmer2017route} in order to optimize the agent routes in a stochastic environment, for-example where RAVs can fail and battery usage may not be fully known. 
\subsection{Image Processing and Scene Analysis}
Our Central Decision Management (CDM) system uses the object labels predicted by a deep neural network from images taken by the RAV cameras. Specifically, we fine-tuned an object detection model \emph{Mask R-CNN} \cite{he2017mask} with our annotated synthetic images that we collected from the virtual scene. Training on a synthetic dataset has been shown to transfer well to real world data in self-driving cars \cite{pan2017virtual} and object detection \cite{tian2018training}. 
\begin{figure}[H]
    \centering
    \includegraphics[width=3.3in]{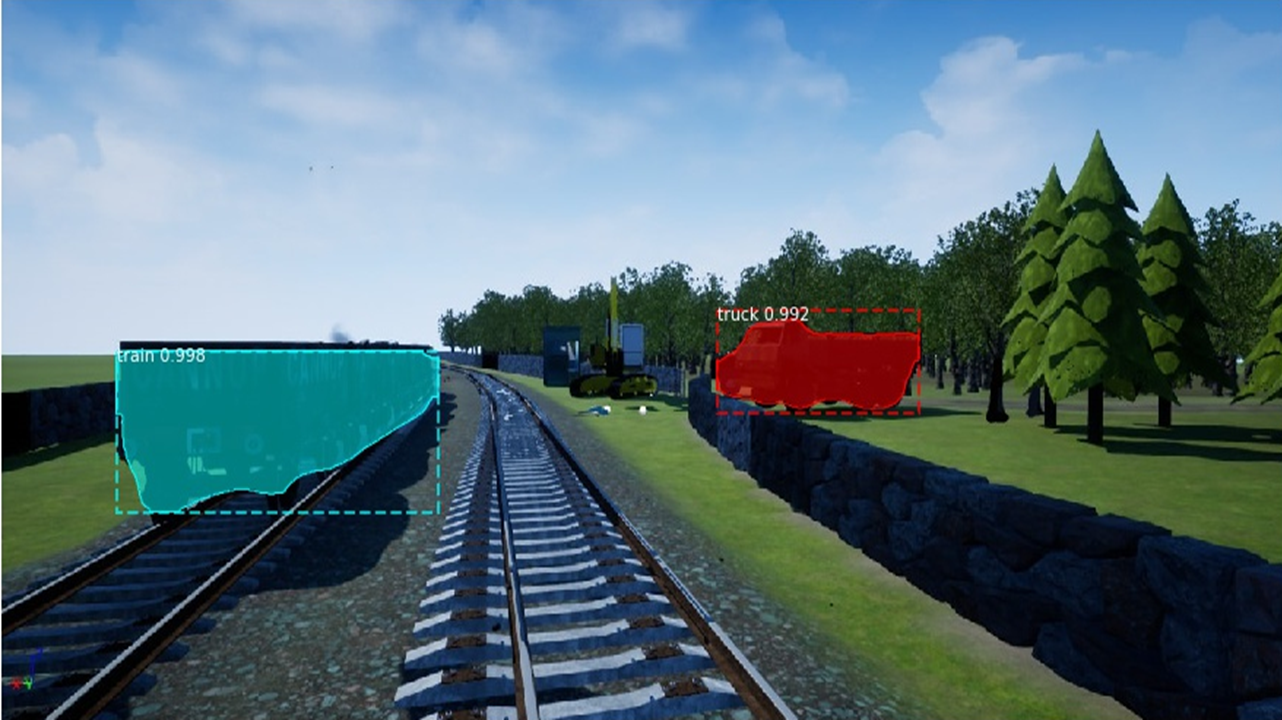}
    \caption{Object identification from a virtual scene image.}
    \vspace{-2em}
    \label{fig:rav}
\end{figure}
\par
\emph{Mask R-CNN} is currently a state-of-the-art object detection deep model that detects and localizes objects with bounding boxes and provides overlay instance segmentation masks to show the contours of the objects within the boxes. Figure \ref{fig:rav} shows the identification of a train and a truck from our virtual scene. The objective of using this detection algorithm is to highlight objects of interest within the scene to the crime scene investigator's attention. These models can detect objects even if they are overlapping. The predicted labels that are produced are an input for our probabilistic reasoning module. Currently, we are enhancing the performance of this deep learning model by retraining/fine-tuning the network on other relevant datasets, for example, Object deTection in Aerial (DOTA) images \cite{xia2017dota}. In addition, our plan is to also detect anomalies in the scenes. 
\subsection{Reasoning and Information Retrieval}
We have developed a probabilistic model in the BLOG language \cite{blog}. It synthesizes data and reasons about the threats in the scene over time. The objective is to estimate the probabilities of different broad categories of threat (chemical, biological, or radiation/nuclear) and specific threat substances. This information affects the way a scene is assessed. For example, a first responder with a hand-held instrument may initially detect evidence of radiation in some regions of the scene. Subsequent RAV images may then show damaged vegetation in those and other regions, which could be caused by radiation
or chemical substances. Another source of information could come from RAVs dispatched with radiation
sensors that fly low over those regions. Using keywords that come from sources such as the object detection module, the probabilistic reasoning module, and the crime scene investigators, the CDM retrieves documentation such as standard operating procedures and guidance documents from a knowledge base. This retrieval is done based on rankings (in order of relevance to the current situation). Elastic Search and a previously-defined set of CBRNE synonyms are used for rankings. The documents are re-ranked in real-time as new information becomes available from various sources.
\bibliographystyle{plain}
\bibliography{ijcai18}





\end{document}